# Uncertainty estimation of machine learning spatial precipitation predictions from satellite data

Georgia Papacharalampous[1,*], Hristos Tyralis[2], Nikolaos Doulamis[3], Anastasios Doulamis[4]

[1] Department of Topography, School of Rural, Surveying and Geoinformatics Engineering, National Technical University of Athens, Iroon Polytechniou 5, 157 80 Zografou, Greece (papacharalampous.georgia@gmail.com, gpapacharalampous@hydro.ntua.gr, https://orcid.org/0000-0001-5446-954X)

[2] Department of Topography, School of Rural, Surveying and Geoinformatics Engineering, National Technical University of Athens, Iroon Polytechniou 5, 157 80 Zografou, Greece (montchrister@gmail.com, hristos@itia.ntua.gr, https://orcid.org/0000-0002-8932-4997)

[3] Department of Topography, School of Rural, Surveying and Geoinformatics Engineering, National Technical University of Athens, Iroon Polytechniou 5, 157 80 Zografou, Greece (ndoulam@cs.ntua.gr, https://orcid.org/0000-0002-4064-8990)

[4] Department of Topography, School of Rural, Surveying and Geoinformatics Engineering, National Technical University of Athens, Iroon Polytechniou 5, 157 80 Zografou, Greece (adoulam@cs.ntua.gr, https://orcid.org/0000-0002-0612-5889)

* Corresponding author

**Abstract**: Merging satellite and gauge data with machine learning produces high-resolution precipitation datasets, but uncertainty estimates are often missing. We addressed the gap of how to optimally provide such estimates by benchmarking six algorithms, mostly novel even for the more general task of quantifying predictive uncertainty in spatial prediction settings. On 15 years of monthly data from over the contiguous United States (CONUS), we compared quantile regression (QR), quantile regression forests (QRF), generalized random forests (GRF), gradient boosting machines (GBM), light gradient boosting machine (LightGBM), and quantile regression neural networks (QRNN). Their ability to issue predictive precipitation quantiles at nine quantile levels (0.025, 0.050, 0.100, 0.250, 0.500, 0.750, 0.900, 0.950, 0.975), approximating the full probability distribution, was evaluated using quantile scoring functions and the quantile scoring rule. Predictors at a site were nearby values from two satellite precipitation retrievals, namely PERSIANN (Precipitation Estimation from Remotely Sensed Information using Artificial Neural Networks) and IMERG (Integrated Multi-satellitE Retrievals), and the site's elevation. The dependent variable was the monthly mean gauge precipitation. With respect to QR, LightGBM showed improved performance in terms of the quantile scoring rule by 11.10%, also surpassing QRF (7.96%), GRF (7.44%), GBM (4.64%) and QRNN (1.73%). Notably, LightGBM outperformed all random forest variants, the current standard in spatial prediction with machine learning. To conclude, we propose a suite of machine learning algorithms for estimating uncertainty in spatial data prediction, supported with a formal evaluation framework based on scoring functions and scoring rules.

## 1. Introduction

Merging satellite-based data and gauge observations can produce accurate precipitation datasets with fine spatial resolution (Hu et al. 2019, Abdollahipour et al. 2022). This is necessary because (a) gauges offer high accuracy, but spatially dense gauge networks are expensive, and (b) satellites provide affordable fine-resolution data but with lower accuracy.

Machine learning algorithms (Hastie et al. 2009, James et al. 2013, Efron and Hastie 2016) often serve as tools for merging satellite-based and gauge precipitation data. This procedure, essentially a regression problem, utilizes gauge measurements as the dependent variable and satellite data as predictor variables. After training a regression algorithm on these paired data points, precipitation can be predicted at any location within the study area and, in this way, a more accurate gridded dataset can be produced (Baez-Villanueva et al. 2020). Relevant works were conducted, for example, by Baez-contiguous United States (CONUS), we compared quantile regression (QR), quantile regression forests (QRF), generalized random forests (GRF), gradient boosting machines (GBM), light gradient boosting machine (LightGBM), and quantile regression neural networks (QRNN). Their ability to issue predictive precipitation quantiles at nine quantile levels (0.025, 0.050, 0.100, 0.250, 0.500, 0.750, 0.900, 0.950, 0.975), approximating the full probability distribution, was evaluated using quantile scoring functions and the quantile scoring rule. Predictors at a site were nearby values from two satellite precipitation retrievals, namely PERSIANN (Precipitation Estimation from Remotely Sensed Information using Artificial Neural Networks) and IMERG (Integrated Multi-satellitE Retrievals), and the site's elevation. The dependent variable was the monthly mean gauge precipitation. With respect to QR, LightGBM showed improved performance in terms of the quantile scoring rule by 11.10%, also surpassing QRF (7.96%), GRF (7.44%), GBM (4.64%) and QRNN (1.73%). Notably, LightGBM outperformed all random forest variants, the current standard in spatial prediction with machine learning. To conclude, we propose a suite of machine learning algorithms for estimating uncertainty in spatial data prediction, supported with a formal evaluation framework based on scoring functions and scoring rules.

## 1. Introduction

Merging satellite-based data and gauge observations can produce accurate precipitation datasets with fine spatial resolution (Hu et al. 2019, Abdollahipour et al. 2022). This is necessary because (a) gauges offer high accuracy, but spatially dense gauge networks are expensive, and (b) satellites provide affordable fine-resolution data but with lower accuracy.

Machine learning algorithms (Hastie et al. 2009, James et al. 2013, Efron and Hastie 2016) often serve as tools for merging satellite-based and gauge precipitation data. This procedure, essentially a regression problem, utilizes gauge measurements as the dependent variable and satellite data as predictor variables. After training a regression algorithm on these paired data points, precipitation can be predicted at any location within the study area and, in this way, a more accurate gridded dataset can be produced (Baez-Villanueva et al. 2020). Relevant works were conducted, for example, by Baez-Villanueva et al. (2020), Chen et al. (2020), Wu et al. (2020), Nguyen et al. (2021), Fernandez-Palomino et al. (2022), Gavahi et al. (2023), and Papacharalampous et al. (2023a), among others.

A characteristic of most of the relevant studies is that they issue point predictions of precipitation at any point in space (Hengl et al. 2018). While this approach corrects the inherent inaccuracies of satellite data using sparse gauge information, the resulting predictions still hold some degree of uncertainty that needs to be quantified. Therefore, entire predictive probability distributions or adequate approximations of them should be preferred when possible (Gneiting and Raftery 2007) due to the larger amount of information that they offer, thereby allowing better decision making than point predictions alone. Such predictive uncertainty estimates (i.e., probabilistic predictions) appear more frequently in other fields (e.g., in Rodrigues and Pereira 2020, Kasraei et al. 2021, Cui et al. 2022) and in other hydrological disciplines (e.g., in Weerts et al. 2011, Tareghian and Rasmussen 2013, Papacharalampous et al. 2020, Tyralis and Papacharalampous 2021b, Tyralis et al. 2023b), but have only been provided by a few studies for the topic of remote sensing of precipitation (Bhuiyan et al. 2018, Zhang et al. 2022, Glawion et al. 2023 and Tyralis et al. 2023a). Still, each of these studies uses a single algorithm and the one by Tyralis et al. (2023a) additionally focuses on extreme quantiles of the predictive probability distribution. Therefore, the benefits that machine learning can bring to the general task of predictive uncertainty quantification in merging satellite and gauge-measured precipitation data have not been sufficiently explored so far.

The aim of this work was to fill in this specific gap for advancing the state of the machine learning-driven applications of predictive uncertainty quantification in merging precipitation observations from satellites and ground-based gauges. This was achieved though extensively investigating the machine learning concept of quantile regression for fulfilling this task. Indeed, as the various quantile regression algorithms exhibit different relative performances at different prediction problems (see diverging results from other fields, for example, in Zhang et al. 2018, He et al. 2019, Sesia and Candès 2020), a comparison between alternative implementations in the field of interest was necessary.

Quantile regression algorithms can provide quantile predictions, allowing them to approximate a predictive probability distribution. Comparisons between quantile regression algorithms rely on quantile scoring (loss) functions and quantile scores, which are well-suited metrics for evaluating quantile predictions. Quantile scoring functions offer a consistent assessment of quantile predictions (Gneiting 2011), while quantile scores serve as proper scoring rules for assessing probabilistic predictions in the form of quantile predictions at multiple quantile levels (Gneiting and Raftery 2007). The formal definitions of consistency and propriety for these metrics are provided in Section 4.3. For now, it suffices to understand that a consistent scoring (loss) function for a point prediction (e.g., a quantile prediction) incentivizes modellers to make attentive and truthful assessments of those predictions (Gneiting 2011). Similarly, a proper scoring rule for a probabilistic prediction encourages modellers to make accurate and honest assessments of the predicted probability distribution (Gneiting and Raftery 2007).

The comparison had to be of large scale according to the guidelines by Boulesteix et al. (2018) to ensure that its output would be as general as possible for the problem of interest. This translates to using data that cover a large spatial region and a large time period, and to including several learners and a large number of predictors in the experiments. Notably, such large-scale comparisons are rare in the literature that merges data from satellites and ground-based gauges (Papacharalampous et al. 2023a).

The remaining article is arranged as follows: Section 2 defines the problem setting. Section 3 provides important information about the learners that were extensively compared by this work for the general task of estimating predictive uncertainty in blending precipitation observations from satellites and ground-based gauges. Section 4 presents the dataset and procedures used for the comparison. Section 5 presents the large-scale results. Section 6 discusses several outputs of the comparison under the spectrum that the current literature provides and proposes ideas for future investigations. Section 7 presents the concluding remarks. To ensure the reproducibility of the methods and tests of this work, Appendix A is devoted to statistical software information. This latter information could be used alongside with the parameter settings described in Section 3.

## 2. Spatial prediction setting and validation method

In this section, we define the problem solved, while detailed descriptions of its components follow in Sections 3 and 4.

## 2.1 Spatial prediction setting

Figure 1 illustrates how the spatial prediction problem was formulated for facilitating the experiments of this work. The gauge-measured total monthly precipitation at a geographical location of interest (e.g., the location of the station represented in Figure 1) was the dependent variable. Grid points are locations of satellite-based precipitation data. Following procedures proposed in previous works (Papacharalampous et al. 2023a, b, c and Tyralis et al. 2023a), the four closest grid points from each grid to each ground-based station were identified and indexed according to their distance from this station, which was computed in meters. The monthly precipitation totals at the four closest grid points 1–4 are hereinafter called "PERSIANN values 1–4" or "IMERG values 1–4", corresponding to the satellite precipitation dataset [PERSIANN (Precipitation Estimation from Remotely Sensed Information using Artificial Neural Networks) and IMERG (Integrated Multi-satellitE Retrievals) are two satellite datasets, that are described in detail later in Section 4.1.2]. Similarly, the distances $d_i$, $i$ = 1, 2, 3, 4, where $d_1 < d_2 < d_3 < d_4$, are hereinafter called "PERSIANN distances 1–4" or "IMERG distances 1–4". These total monthly precipitation values and distances were the predictor variables, together with the elevation at the station. Practically, the spatial prediction setting is a regression problem.

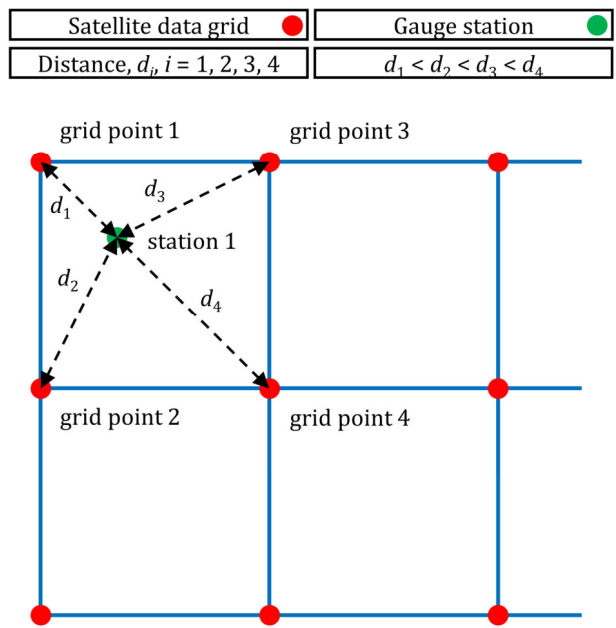

Figure 1. Illustration of how the spatial prediction problem was formulated. The satellite precipitation data are available at grid points. The distances of a given station (station 1) from its four closest grid points (grid points 1–4) are denoted with $d_i$, $i$ = 1, 2, 3 and 4. These distances and the monthly precipitation data from satellites at the same grid points were used as predictors for precipitation at the given station.

The data of the regression problem are described in more detail later:

- Description and location of stations in Section 4.1.1.
- Description and location of satellite data (PERSIANN and IMERG data) in Section 4.1.2.
- Computation of stations' elevation in Section 4.1.3.

The dataset encompassed 91 623 samples spanning the contiguous United States (CONUS) from 2001 to 2015. These samples were utilized within a five-fold cross-validation setting. Each sample included a monthly precipitation measurement from a station (dependent variable), along with the corresponding monthly precipitation values from the four closest grid points in the PERSIANN dataset and their respective distances to the station (8 predictor variables), as well as 8 predictor variables from the IMERG dataset and the station's elevation (1 predictor variable). In total, each sample included the dependent variable value and the values of 17 predictor variables.

## 2.2 Predictions at spatial points in the form of quantiles

In a regression setting, a trained quantile regression algorithm can issue a prediction of a quantile $Q_{\underline{z}}(\alpha)$ of the probability distribution of a predicted random variable $\underline{z}$ at a pre-specified level $\alpha$, where $Q_{\underline{z}}(\alpha)$ is defined by:

$$Q_{\underline{z}}(\alpha) := F_{\underline{z}}^{-1}(\alpha), 0 \leq \alpha \leq 1, \qquad (1)$$

Assume that one is interested in predicting the median of the prediction, i.e., the $Q_{\underline{z}}(0.5)$ quantile. Then, the assessment of two algorithms that can predict medians can be done with the absolute error scoring function averaged over the test set (i.e., using the mean absolute error, MAE).

Returning to our problem, let us assume that we receive the directive to predict monthly precipitation quantiles at multiple levels other than 0.500 at a point in space. By doing so, it is possible to approximate the probability distribution of the prediction at any point, because we can compute the predictor variables (satellite precipitation of four closest points, distances of the four points and the point's elevation) at any point in space. Given that quantile prediction can be done using multiple quantile regression algorithms, the natural question is which algorithm to choose. That requires a formal assessment which is based on three components:

- Existence of ground-based measurements for training and testing the algorithms.
- Use of quantile scoring functions as metrics for the assessment of quantile predictions.

Quantile scoring functions are generalizations of the absolute error scoring function and are suitable for assessing quantile predictions (details on quantile scoring functions can be found in Section 4.3).

- If the full probabilistic predictions need to be assessed, then the comparison should be based on quantile scores that are proper scoring rules, i.e., suitable to assess probabilistic predictions in the form of multiple quantiles (details on quantile scores can be found in Section 4.3).

Here, it is important to note that:

- The proposed framework does not study the problem of the discrepancies between satellite-based datasets and gauge-based measurements. It assumes that gauge-based measurements are correct (free of measurement errors) and represent the ground truth.

- It estimates a probability distribution of a prediction, in the sense that predictions should be provided in the form of a probability distribution.

- Since multiple algorithms can do this, we can select the most appropriate based on a metric. In our case, the implemented metrics were quantile scoring functions and quantile scoring rules.

- If one algorithm outperforms the others in the test set, it is natural to prefer this algorithm for predictions at any point in space for the specific dataset.

## 2.3  Problem definition

Based on the previous descriptions, the problem was defined as follows:

a. Let $PR_{i,j}$ denote the monthly precipitation at the gauge $i$ in the time point (i.e., month) $j$.

b. Compute the elevation $EL_i$ of the gauge $i$.

c. Compute the distances of the closest satellite grid points from the gauge $i$ ($d_{1,PERSIANN}$, $d_{2,PERSIANN}$, $d_{3,PERSIANN}$, $d_{4,PERSIANN}$, and $d_{1,IMERG}$, $d_{2,IMERG}$, $d_{3,IMERG}$, $d_{4,IMERG}$).

d. Compute the monthly precipitation of the closest satellite grid points from the gauge $i$ in the same time point $j$ ($PR_{1,PERSIANN}$, $PR_{2,PERSIANN}$, $PR_{3,PERSIANN}$, $PR_{4,PERSIANN}$, and $PR_{1,IMERG}$, $PR_{2,IMERG}$, $PR_{3,IMERG}$, $PR_{4,IMERG}$).

e. Form 91 623 samples in the form $SAMPLE_{i,j}$ = {$PR_{i,j}$, $EL_i$, $d_{1,PERSIANN}$, $d_{2,PERSIANN}$, $d_{3,PERSIANN}$, $d_{4,PERSIANN}$, $d_{1,IMERG}$, $d_{2,IMERG}$, $d_{3,IMERG}$, $d_{4,IMERG}$, $PR_{1,PERSIANN}$, $PR_{2,PERSIANN}$, $PR_{3,PERSIANN}$, $PR_{4,PERSIANN}$, $PR_{1,IMERG}$, $PR_{2,IMERG}$, $PR_{3,IMERG}$, $PR_{4,IMERG}$}.

f. Define a quantile regression problem at quantile level $α$, where $PR_{i,j}$ is the dependent

variable and the remaining 17 variables of SAMPLE$_{i,j}$ are the predictor variables.

g. Test multiple quantile regression algorithms in a five-fold cross validation setting using the 91 623 samples.

h. The best algorithm, according to the pre-specified metric (quantile scoring function or quantile score), can be retrained in the full sample and, then, it will be able to predict quantiles at level $α$, at any point in space. Please note that those predictions are quantiles of the probability distribution of the prediction. Consequently, if the procedure is repeated for multiple quantile levels, we will have an approximation of the full probability distribution of the prediction at any point.

## 3. Learners for uncertainty estimation

We recall from Section 2 that the dataset comprised 91 623 samples, with each sample encompassing the values of the dependent variable and 17 predictor variables, and that the assessment is a five-fold cross-validation setting. The six quantile regression algorithms described in this section were assessed in the study. Recall also that, once a quantile regression algorithm is trained, then it can predict quantiles at multiple pre-specified quantile levels. Default software implementation parameters (where applicable) were selected. It is well known that random forests (and their variants) perform well with default hyperparameter values. For the remaining algorithms (boosting and neural networks), optimization to force performance to the limits was out of scope, while those algorithms perform also well with default software hyperparameters.

### 3.1 Quantile regression

Quantile regression (QR; Koenker and Bassett 1978, Koenker 2005) is the simplest learner used in this work (as it is linear) for estimating predictive uncertainty in blending precipitation data from satellites and ground-based gauges, and is trained by minimizing across the various samples the quantile scoring function (Gneiting 2011). The latter is defined in Section 4.3. Tutorial information about QR can be found in Waldmann (2018).

### 3.2 Quantile regression forests

Quantile regression forests (QRF; Meinshausen and Ridgeway 2006) are a variant of random forests (Breiman 2001) for predictive uncertainty quantification. Random forests fall into the category of ensemble learning algorithms (Sagi and Rokach 2018). They grow an ensemble of decision trees (Breiman et al. 1984) by running bootstrap aggregation (bagging; Breiman 1996) coupled with an additional randomization procedure. While, in standard regression, the output prediction is the mean of the predictions of all the decision trees (with this mean approximating the conditional mean of the response variable), for QRF it is a resemblance of the conditional distribution of the response variable. In this work, QRF were implemented with 500 decision trees. The other parameters of QRF were set to their defaults in Tibshirani and Athey (2023).

### 3.3 Generalized random forests

Generalized random forests (GRF; Athey et al. 2019) are another variant of random forests for uncertainty quantification that are expected to model heterogeneities better than QRF. In this work, GRF were implemented with 500 decision trees. The other parameters of GRF were set to their defaults in Tibshirani and Athey (2023).

### 3.4 Gradient boosting machines

Gradient boosting machines (GBM; Friedman 2001) are a boosting (Mayr et al. 2014, Tyralis and Papacharalampous 2021a) variant that can be used, among others, for predictive uncertainty quantification. A boosting algorithm is an ensemble learning algorithm that composes a strong learner by sequentially adding weak base learners to the ensemble. Specifically, the training of every new base learner targets at minimizing the error of the current ensemble. This error is the quantile loss (Gneiting 2011) for the case of GBM. Notably, the number of times that a new base learner is added to the ensemble (which is the same with the number of base learners) should be that large that it ensures proper fitting, simultaneously avoiding overfitting. In this work, GBM employed 500 decision trees as their base learners. The other parameters of GBM were set to their defaults in Greenwell et al. (2022).

### 3.5 Light gradient boosting machine

Light gradient boosting machine (LightGBM; Ke et al. 2017) is the second boosting variant used in this work. It is known to have some better properties (e.g., smaller training time) compared to GBM (Tyralis and Papacharalampous 2021a). In this work, it was implemented with decision trees as base learners. The maximum tree depth, the maximum number of leaves in one tree, the minimal number of data in one leaf, the shrinkage rate, the number of boosting iterations (number of trees), the feature fraction that is randomly selected on each iteration (tree), the data fraction that is randomly selected without resampling on each iteration (tree), the minimal gain to perform split and the number of threads were equal to 10, 500, 200, 0.05, 400, 0.75, 0.75, 0 and 10, respectively. These are parameter values that were previously identified as optimal in Tyralis et al. (2023a) for a similar problem. The other parameters of LightGBM were set to their defaults in Shi et al. (2023).

## 3.6 Quantile regression neural networks

Quantile regression neural networks (QRNN; Taylor 2000, Cannon 2011) are a neural network (Hastie et al. 2009, chapter 11) algorithm that can, by its construction, facilitate uncertainty quantification by minimizing a quantile scoring function. In modelling with neural networks, linear combinations of the predictors are first defined. Then, the form of a non-linear function of these combinations is given to the dependent variable. In this work, QRNN were run with a number of repeated trials equal to 1. The other parameters of QRNN were set to their defaults in Cannon (2023).

## 4. Data, application and comparison

### 4.1 Data

To evaluate and compare the learners, three types of data were used. These were gauge-measured precipitation (see Section 4.1.1), satellite precipitation (see Section 4.1.2) and elevation (see Section 4.1.3) data. Note that the same data compilation was previously used for different benchmarking purposes in Papacharalampous et al. (2023c).

#### 4.1.1 Gauge-measured precipitation data

Monthly precipitation totals from ground-based gauges were extracted from the Global Historical Climatology Network monthly database, version 2 (GHCNm) (Peterson and Vose 1997). The extraction took place through the repository of the National Oceanic and Atmospheric Administration (NOAA), which is available online in the following address (accessed on 2022-09-24): https://www.ncei.noaa.gov/pub/data/ghcn/v2. The gauge-measured data refer to 1 421 locations (see Figure 2) in the CONUS and to the time period starting from 2001 and ending with 2015.

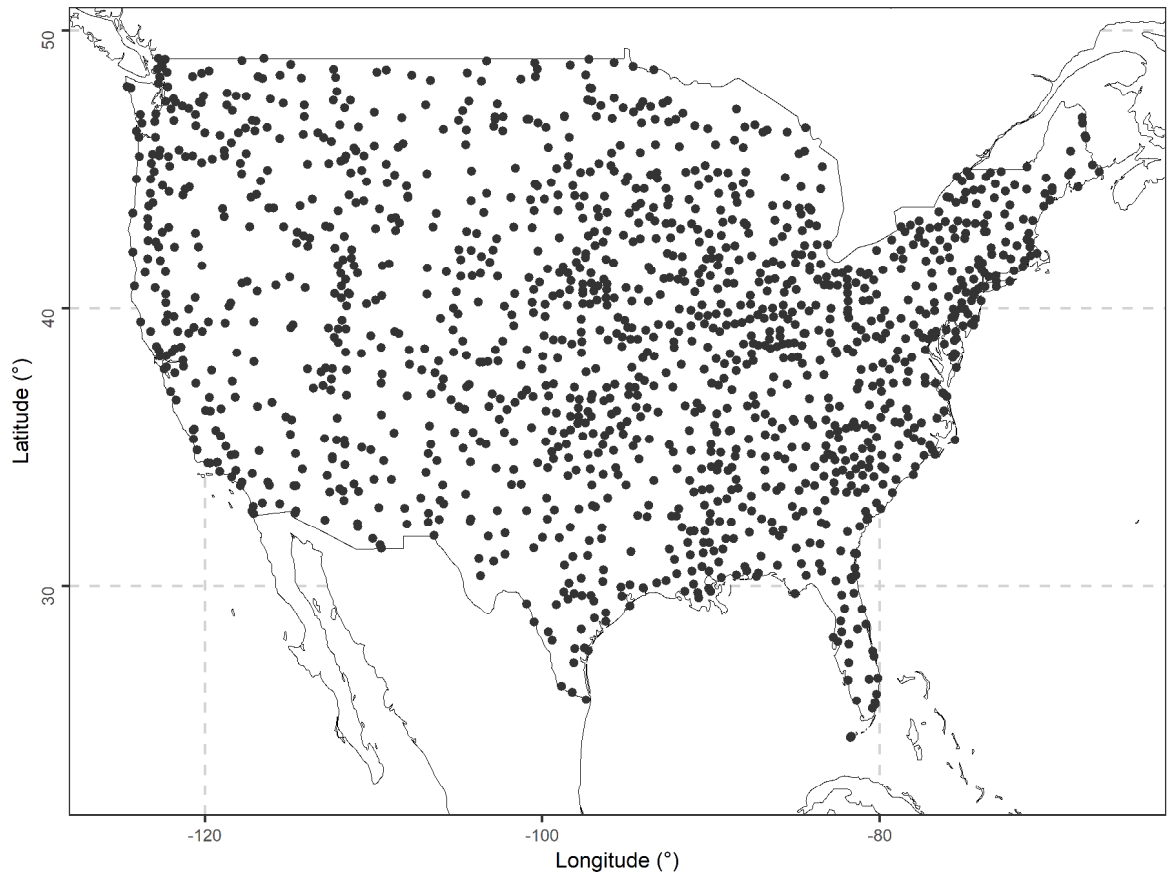

Figure 2. Locations of the ground-based gauges that recorded data used in this study.

*4.1.2   Satellite precipitation data*

Precipitation data from satellites were extracted from two databases, specifically the current operational PERSIANN database (Hsu et al. 1997, Nguyen et al. 2018, 2019) and the GPM IMERG late Precipitation L3 1 day 0.1 degree x 0.1 degree V06 database (Huffman et al. 2019). These databases were formed by the Centre for Hydrometeorology and Remote Sensing (CHRS) of the University of California, Irvine (UCI) and the NASA (National Aeronautics and Space Administration) Goddard Earth Sciences (GES) Data and Information Services Center (DISC), respectively.

In particular, the extraction of gridded satellite data was conducted from the CHRS and the NASA Earth Data repositories, which are available online in the following addresses (accessed on 2022-03-07 and 2022-12-10, respectively): https://chrsdata.eng.uci.edu and https://doi.org/10.5067/GPM/IMERGDL/DAY/06. The satellite data extracted were daily precipitation data and referred to the time period starting from 2001 and ending with 2015. The PERSIANN and IMERG grids extracted cover the CONUS entirely and have a spatial resolution of 0.25 degree x 0.25 degree and 0.1 degree x 0.1 degree, respectively. Bilinear interpolation was applied to the IMERG grid extracted to transform it to the CMORPH0.25 grid with a spatial resolution of 0.25 degree x 0.25 degree. The latter IMERG grid was used in the experiments of this work, together with the PERSIANN grid extracted. Additionally, total monthly precipitation data was obtained by aggregating the daily data for conforming to requirements of the same experiments.

*4.1.3   Elevation data*

Because of its well-recognised usefulness as a predictor for many hydrometeorological variables (Xiong et al. 2022), elevation was estimated at the locations of the ground-based gauges (which are depicted in Figure 2). The estimation was based on the Amazon Web Services (AWS) Terrain Tiles. The latter are available online in the following address (accessed on 2022-09-25): https://registry.opendata.aws/terrain-tiles.

4.2   Quantile levels and quantile crossing handling

A good approximation of the predictive probability distribution can be acquired by predicting quantiles of this distribution at multiple levels. In this study, the quantile levels $\alpha \in \{0.025, 0.050, 0.100, 0.250, 0.500, 0.750, 0.900, 0.950, 0.975\}$ were investigated. We preferred to predict quantiles at the tails and, consequently, most of the results refer to the ability of the algorithms to model the probability mass at the tails of the probability density function. Results could be different if we were interested on the central mass of the probability density function. Negative predicted quantiles at the lowest quantile level were censored to zero. Additionally, quantile crossing was handled by setting, separately for each pair {sample, learner}, any predictive quantile that was smaller than the predictive quantile of the immediate lower quantile level equal to this latter predictive quantile.

4.3   Performance comparison

To facilitate the comparison of the six learners (see Section 3) with respect to their performance at the nine quantile levels (see Section 4.2), the quantile scoring function was first computed. This function is defined as

$$L_\alpha(z, y) := f_\alpha(z - y), \quad (2)$$

where $y$ is the realization of the precipitation process, $z$ is the respective prediction (predictive quantile) at level $\alpha$ and $f_\alpha(x)$ is defined as

$$f_\alpha(x) := x(\mathbb{I}(x \geq 0) - \alpha), x \in \mathbb{R}, \quad (3)$$

where $\mathbb{I}(A)$ is the indicator function. This function is equal to 1 when the event $A$ realizes.

Otherwise, it is equal to 0.

The quantile scoring function is asymmetric around the realization $y$, thereby allowing to predict quantiles (an illustration of the quantile scoring function for the case of $y = 0$ can be found in Tyralis and Papacharalampous 2024). Setting $\alpha = 1/2$ in Equation (2), we get $L_{1/2}(z, y)$ which is the absolute error scoring function up to a multiplicative constant.

$$L_{1/2}(z, y) := |z - y|/2 \tag{4}$$

The absolute error scoring function is symmetric with respect to the realization $y$ and can be used to rank predictions of the median. The quantile scoring function is strictly consistent for the quantile at level $\alpha$ (Gneiting 2011), i.e., it has the following property:

$$\mathbb{E}_F[L(q, \underline{y})] \leq \mathbb{E}_F[L(z, \underline{y})] \; \forall \; F \in \mathcal{F}, q \in Q(F), z \in \mathbb{R}, \tag{5}$$

where $\mathcal{F}$ is the family of probability distributions $F$ of the random variable $\underline{y}$ and $q$ is the actual value of the predictive quantile. Strict consistency for the quantile scoring functions means that its expectation is minimized when the prediction $z$ attains the true quantile value $q$.

A performance criterion takes the form

$$\bar{L}_\alpha(z, y) := (1/k) \sum_{i=1}^{k} L_\alpha(z_i, y_i), \tag{6}$$

where $z_i$ and $y_i$, $i \in \{1, \ldots, k\}$ are, respectively, the prediction and observation of the $i^{\text{th}}$ data sample and $k$ is the number of data samples included in the test dataset. Obviously, this performance criterion was computed separately for each learning algorithm.

Quantile prediction skills $\bar{L}_{\alpha,\text{skill}}$ were also computed for each learning algorithm and the entire dataset according to the equation

$$\bar{L}_{\alpha,\text{skill}} := 1 - \bar{L}_{\alpha,\text{learner}}/\bar{L}_{\alpha,\text{benchmark}}, \tag{7}$$

where the simplest learner (i.e., QR) is used as benchmark. The quantile prediction skill takes values between $-\infty$ and 1. The larger it is, the better the predictive quantiles of a learner on average compared to the predictive quantiles of the benchmark. The benchmark method and the learner are equivalent with respect to their performance when $\bar{L}_{\alpha,\text{skill}} = 0$.

Quantile prediction skills were additionally computed for each pair {learner, station} to examine how the relative performance of the learners varies from station to station for each quantile level.

Moreover, the quantile scoring functions at the various quantile levels (computed as stated above using Equation (2)) were summed up for each pair {case, learner}. The sum of quantile scoring functions for quantile predictions $z_1, \ldots, z_k$ at levels $\alpha_1, \ldots, \alpha_k$ defines the quantile scoring rule $S$:

$$S(z_1, \ldots, z_k; y) := \sum_{i=1}^{k} L_{\alpha_i}(z_i, y), \tag{8}$$

which is proper (Gneiting and Raftery 2007). In particular, the expectation of a proper scoring rule is minimized when the prediction of the probability distribution attains its true value. Thus, proper scoring rules incentivize modellers to report the true predictive probability distribution.

The values of the quantile scoring rule at each point acquired through this procedure were subsequently averaged across all the cases (points) in the dataset, separately for each learner, thereby giving average scores. Then, following an equation similar to Equation (7), the latter scores were used to compute quantile scoring rule skills (Gneiting and Raftery 2007). Similar to the quantile prediction skill, the quantile scoring rule skill takes values from $-\infty$ to 1 and the larger it is, the better the predictive performance with respect to the benchmark. Quantile scoring rule skills were also computed for each pair {learner, station} to examine how the relative performance of the learners varies from station to station for the entire approximation of the predictive probability distribution.

Lastly, the reliability of the learners was examined by computing sample coverages. For this test, the frequency with which each predictive quantile is larger or equal to its corresponding observation was computed for each pair {learner, quantile level}. The closer a frequency to its nominal value, the more reliable the predictive quantiles.

### 4.4 Predictor variable importance comparison

Explainable machine learning can support, among others, predictor variable importance comparisons. For conducting such a comparison for the problem of interest in this study, LightGBM (see Section 3.5) was used. More precisely, the total gain in splits of each predictor variable (Shi et al. 2023) was computed through this learner for the entire dataset. Based on the respective results, the predictors were ranked from the most to the least important ones according to the following rule: The larger the total gain, the larger the importance.

# 5. Results

## 5.1 Performance comparison

Figure 3 presents statistics that allow us to evaluate and compare the performance of the learners in the task of estimating predictive uncertainty while blending precipitation data from satellites and ground-based gauges. In terms of sample coverage (see Figure 3a), all the learners performed at least adequately well in the large-scale experiment, with the best of them being QRF and GRF for quantile levels {0.025, 0.050, 0.100, 0.250}, QR and QRNN for the quantile level 0.500, and QR, GBM and QRNN for the remaining quantile levels.

The goal of probabilistic prediction is to maximize sharpness subject to calibration (Gneiting and Raftery 2007). Nice coverage can indicate nice calibration, thereby offering some insight into the absolute performance of the learners. Given that all algorithms seem well calibrated (i.e., they have good coverages), the one that maximized sharpness should be preferred. Sharpness is a property of the predictions. As a representative intuitive example, sharpness of prediction intervals can be characterized by their width, where narrower intervals are preferable. Scoring rules provide a single metric encompassing both calibration and sharpness, enabling the honest ranking of probabilistic predictions.

According to the quantile prediction skill (see Figure 3b), the best-performing learner is LightGBM at all quantile levels. According to the same skill, QRF, GRF, GBM, QRNN and QR are, respectively, ranked second, third, fourth, fifth and sixth for each of the quantile levels. The same ranking is obtained based on the quantile prediction skill (see Figure 3c), which is appropriate for evaluating the quality of an ensemble of predictions at multiple quantile levels (with these predictions approximating an entire predictive probability distribution). Compared to QR, LightGBM showed improved performance in terms of the quantile scoring rule by 11.10%, surpassing QRF (7.96%), GRF (7.44%), GBM (4.64%), and QRNN (1.73%).

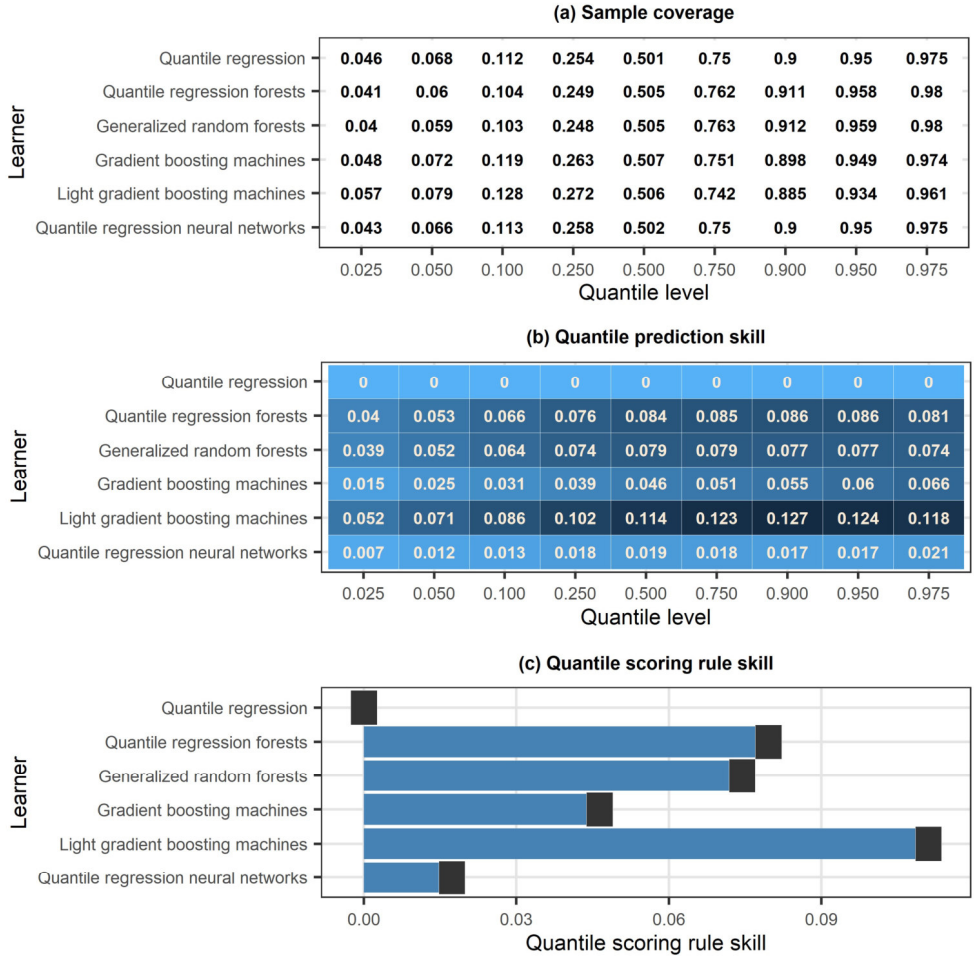

Figure 3. Statistics for facilitating the evaluation and comparison of the learners based on the entire dataset. Specifically: (a) Sample coverage of the learners at the various quantile levels; (b) quantile prediction skill of the learners at the various quantile levels; and (c) quantile scoring rule skill of the learners. For interpreting (a): The closest the sample coverage to the quantile level, the more reliable the quantile predictions. For interpreting (b): The larger the quantile prediction skill (and the darker the colour on the respective heatmap), the better the quantile predictions on average compared to the quantile predictions of the benchmark learner (i.e., quantile regression). For interpreting (c): The larger the quantile scoring rule skill, the better the probabilistic predictions on average compared to the probabilistic predictions of the benchmark learner (i.e., quantile regression).

Figure 4 allows us to assess to which degree the quantile prediction skill can vary in space for the various learners and the various quantile levels, while Figure 5 allows a similar assessment but for the quantile scoring rule skill (which provides a proper assessment for the total of the quantile levels). Both scores were found to vary a lot from station to station. Also notably, good and bad performances with respect to the benchmark tended to appear at the same stations for multiple methods. Therefore, although on average LightGBM outperformed the other algorithms, that does not imply that it was uniformly best, i.e., the best at all individual stations.

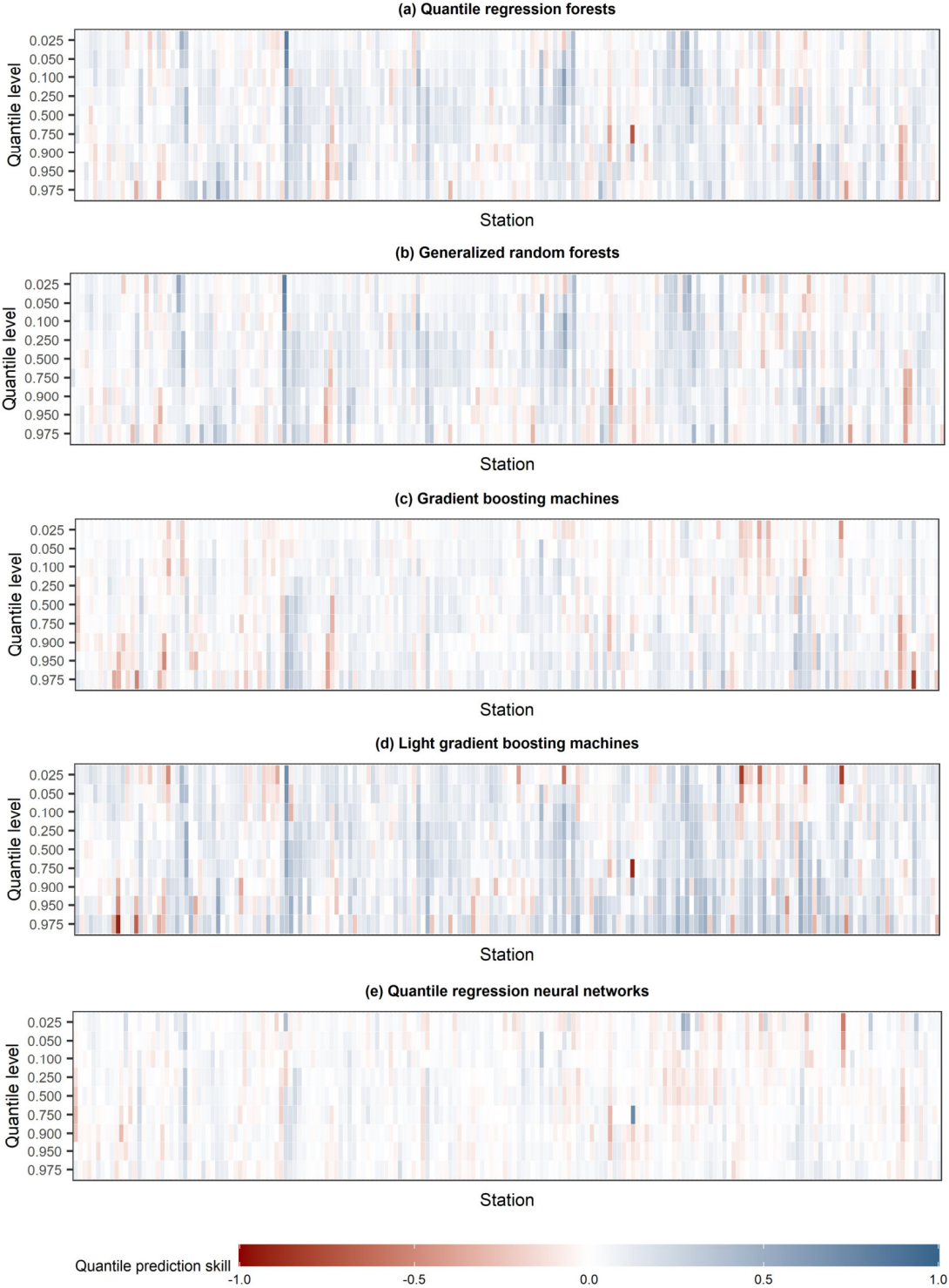

Figure 4. Quantile prediction skill of (a) quantile regression forests, (b) generalized random forests, (c) gradient boosting machines, (d) light gradient boosting machine, and (e) quantile regression neural networks at the various quantile levels for 200 arbitrary stations. The larger the quantile prediction skill (and the darker blue the colour on the heatmaps), the better the quantile predictions on average compared to the quantile predictions of the benchmark learner (i.e., quantile regression).

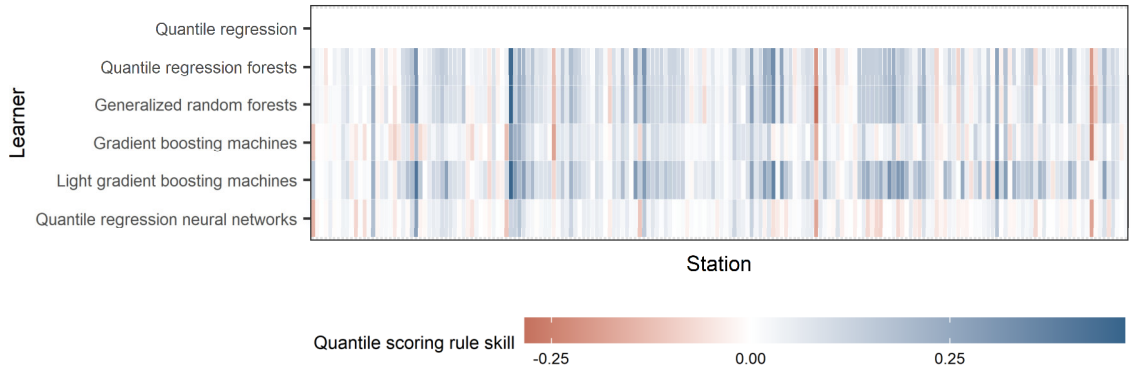

Figure 5. Quantile scoring rule skill of the learners for the same stations as in Figure 4. The larger the quantile scoring rule skill (and the darker blue the colour on the heatmap), the better the probabilistic predictions on average compared to the probabilistic predictions of the benchmark learner (i.e., quantile regression).

## 5.2   Predictor variable importance comparison

Figure 6 facilitates a comparison of the predictors with respect to their importance for the various quantile levels in the task of estimating predictive uncertainty while blending precipitation data from satellites and ground-based gauges. The respective results are largely homogenous across quantile levels, with IMERG value 1 and 2 having been identified as the most and the second most important predictors by the LightGBM algorithm. The remaining IMERG values and the PERSIANN values were most often ranked before the distances with few exceptions, and the elevation of the station is also ranked third or fourth for the quantile levels {0.250, 0.500, 0.750, 0.900, 0.950, 0.975} and eighth or ninth for the remaining quantile levels.

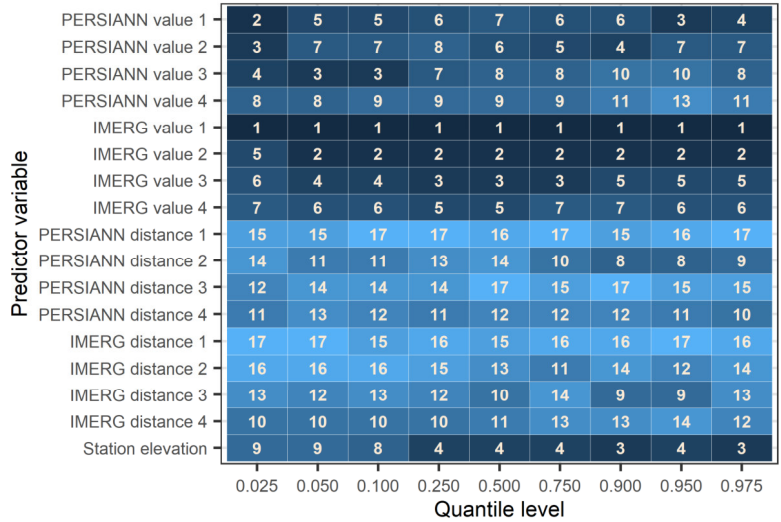

Figure 6. Ranking of the predictors from the most (1st) to the least (17th) important ones at the various quantile levels based on the gain statistic computed through the light gradient boosting machine algorithm by exploiting the entire dataset. The darker the colour on the heatmap, the more important the predictor variable.

## 6. Discussion

The large-scale benchmark experiment of this work suggests that LightGBM is the best-performing learner for uncertainty quantification in blending precipitation data from satellites and ground-based gauges at the monthly temporal scale, leaving QRF, GRF, GBM, QRNN and QR behind, for the case of CONUS. In our view, this is an important finding, given also that LightGBM has at least similar or considerably lower computation requirements than other sophisticated learners for uncertainty quantification and has been proven useful when the interest is in extreme events (Tyralis et al. 2023a). Existing research favours random forests for spatial prediction, but we expect a shift towards boosting methods.

Most algorithms outperformed the linear QR, likely due to their increased flexibility, which leads to better performance on large datasets. A potential drawback limiting the performance of QRNN is the redundancy of predictor variables, particularly the repeated distance variables throughout the sample. This redundancy can be largely mitigated by using tree-based algorithms like QRF and GRF. However, these methods face limitations in extrapolation beyond the training data and tend to focus on the central region of the predictive density function. Consequently, the ability of LightGBM to handle these aspects naturally explains its superior performance. Notably, the results might differ in different spatial settings or with smaller datasets, where simpler methods like linear approaches may be more favourable.

Furthermore, the feature importance tests showed that satellite precipitation variables are more important predictors than the elevation at a point of interest, which is a more important predictor than distances between a point of interest and satellite grid points. The same tests indicated that IMERG provides more useful precipitation information compared to PERSIANN. This outcome agrees with results by Papacharalampous et al. (2023c), a study on satellite precipitation product correction that did not involve uncertainty quantification.

Future research could focus on potential benefits that could stem from the application of the ensemble learning concept (Wang et al. 2022) for dealing with the general task of interest. Such investigations could focus both on simple ways to combine algorithms (e.g., the one in Petropoulos and Svetunkov 2020) and on advanced stacked generalization procedures (e.g., those in Wolpert 1992), and could use the learners implemented and compared in this work as base learners in the ensemble learning. They could also benefit from the comparative framework of this work.

Another idea for future research is to extend the comparison to other families of machine learning algorithms for predictive uncertainty quantification. Summaries of such families and their known properties are available in the reviews by Papacharalampous and Tyralis (2022) and Tyralis and Papacharalampous (2023). Lastly, potential benefits from incorporating time series features, such as those computed in Kang et al. (2017), into the predictive uncertainty quantification frameworks in the field of satellite precipitation product correction could be investigated.

## 7. Summary and conclusions

This is the first general study on the use of machine learning for the important, yet rarely performed, task of estimating predictive uncertainty in merging precipitation data from satellites and ground-based gauges. The concept of quantile regression was extensively investigated for the first time for fulfilling this task. To provide useful guidance on how this concept can be exploited, a large-scale comparison was conducted between quantile regression (QR), quantile regression forests (QRF), generalized random forests (GRF), gradient boosting machines (GBM), light gradient boosting machine (LightGBM) and quantile regression neural networks (QRNN). The comparison took place at the monthly temporal scale using 15-year-long data that span across the contiguous United States.

Such a large-scale benchmark study was indeed necessary, as the relative performance of the various learners can vary substantially from problem to problem and can only be assessed by talking advantage of large datasets. In the context of interest, LightGBM was identified as the best-performing learner for predictive uncertainty estimation based on the quantile prediction and quantile scoring rule skills. The order from the best to the worst of the remaining algorithms is the following: QRF, GRF, GBM, QRNN and QR.

Lastly, the concept of feature importance was exploited through LightGBM to rank the predictors from the most to the least important ones for predictive uncertainty estimation in blending monthly precipitation data from satellites and ground-based gauges, thereby ensuring an adequate degree of interpretability. Among the PERSIANN (Precipitation Estimation from Remotely Sensed Information using Artificial Neural Networks) and IMERG (Integrated Multi-satellitE Retrievals) satellite datasets, the latter consistently improved more the precipitation predictions for most of the quantile levels compared to the former.

**Conflicts of interest:** There is no conflict of interest.

**Funding:** This work was conducted in the context of the research project BETTER RAIN (BEnefiTTing from machine lEarning algoRithms and concepts for correcting satellite RAINfall products). This research project was supported by the Hellenic Foundation for Research and Innovation (H.F.R.I.) under the "3rd Call for H.F.R.I. Research Projects to support Post-Doctoral Researchers" (Project Number: 7368).

**Acknowledgements:** We thank the Editor and the Associate Editor for handling the review process, and the Reviewers whose critical comments resulted in a substantially improved version of the manuscript.

**Data availability statement:** No new data were created or analysed in this study. The datasets used are cited in the manuscript.

**Ethical statement:** Not applicable.

## Appendix A  Statistical software

The R programming language (R Core Team 2023) was used for processing the data, for implementing the learners, and for reporting and visualizing the results. The following R packages were used: `caret` (Kuhn 2023), `data.table` (Dowle and Srinivasan 2023), `devtools` (Wickham et al. 2022), `elevatr` (Hollister 2023), `gbm` (Greenwell et al. 2022), `grf` (Tibshirani and Athey 2023), `knitr` (Xie 2014, 2015, 2023), `lightgbm` (Shi et al. 2023), `ncdf4` (Pierce 2023), `qrnn` (Cannon 2011, 2018, 2023), `quantreg` (Koenker 2023), `rgdal` (Bivand et al. 2023), `rmarkdown` (Allaire et al. 2023, Xie et al. 2018, 2020), `scoringfunctions` (Tyralis and Papacharalampous 2023, 2024), `sf` (Pebesma 2018, 2023), `spdep` (Bivand 2023, Bivand and Wong 2018, Bivand et al. 2013) and `tidyverse` (Wickham et al. 2019, Wickham 2023).

## References

[1] Abdollahipour A, Ahmadi H, Aminnejad B (2022) A review of downscaling methods of satellite-based precipitation estimates. Earth Science Informatics 15(1):1–20. https://doi.org/10.1007/s12145-021-00669-4.

[2] Allaire JJ, Xie Y, McPherson J, Luraschi J, Ushey K, Atkins A, Wickham H, Cheng J, Chang W, Iannone R (2023) rmarkdown: Dynamic Documents for R. R package version 2.21. https://CRAN.R-project.org/package=rmarkdown.

[3] Athey S, Tibshirani J, Wager S (2019) Generalized random forests. Annals of Statistics 47(2):1148–1178. https://doi.org/10.1214/18-AOS1709.

[4] Baez-Villanueva OM, Zambrano-Bigiarini M, Beck HE, McNamara I, Ribbe L, Nauditt A, Birkel C, Verbist K, Giraldo-Osorio JD, Xuan Thinh N (2020) RF-MEP: A novel random forest method for merging gridded precipitation products and ground-based measurements. Remote Sensing of Environment 239:111606. https://doi.org/10.1016/j.rse.2019.111606.

[5] Bhuiyan MAE, Nikolopoulos EI, Anagnostou EN, Quintana-Seguí P, Barella-Ortiz A (2018) A nonparametric statistical technique for combining global precipitation datasets: Development and hydrological evaluation over the Iberian Peninsula. Hydrology and Earth System Sciences 22(2):1371–1389. https://doi.org/10.5194/hess-22-1371-2018.

[6] Bivand RS (2023) spdep: Spatial Dependence: Weighting Schemes, Statistics. R package version 1.2-8. https://CRAN.R-project.org/package=spdep.

[7] Bivand RS, Wong DWS (2018) Comparing implementations of global and local indicators of spatial association. TEST 27(3):716–748. https://doi.org/10.1007/s11749-018-0599-x.

[8] Bivand RS, Pebesma E, Gómez-Rubio V (2013) Applied Spatial Data Analysis with R. Second Edition. Springer New York, NY. https://doi.org/10.1007/978-1-4614-7618-4.

[9] Bivand RS, Keitt T, Rowlingson B (2023) rgdal: Bindings for the 'Geospatial' Data Abstraction Library. R package version 1.6-6. https://CRAN.R-project.org/package=rgdal.

[10] Boulesteix AL, Binder H, Abrahamowicz M, Sauerbrei W (2018) Simulation Panel of the STRATOS Initiative. On the necessity and design of studies comparing statistical methods. Biometrical Journal 60(1):216–218. https://doi.org/10.1002/bimj.201700129.

[11] Breiman L (1996) Bagging predictors. Machine Learning 24(2):123–140. https://doi.org/10.1007/BF00058655.

[12] Breiman L (2001) Random forests. Machine Learning 45(1):5–32. https://doi.org/10.1023/A:1010933404324.

[13] Breiman L, Friedman J, Stone CJ, Olshen RA (1984) Classification and Regression Trees. First Edition. Chapman & Hall/CRC: Boca Raton, FL, USA.

[14] Cannon AJ (2011) Quantile regression neural networks: Implementation in R and application to precipitation downscaling. Computers and Geosciences 37(9):1277–1284. https://doi.org/10.1016/j.cageo.2010.07.005.

[15] Cannon AJ (2018) Non-crossing nonlinear regression quantiles by monotone composite quantile regression neural network, with application to rainfall extremes. Stochastic Environmental Research and Risk Assessment 32(11):3207–3225. https://doi.org/10.1007/s00477-018-1573-6.

[16] Cannon AJ (2023) qrnn: Quantile Regression Neural Network. R package version 2.1. https://CRAN.R-project.org/package=qrnn.

[17] Chen H, Chandrasekar V, Cifelli R, Xie P (2020) A machine learning system for precipitation estimation using satellite and ground radar network observations. IEEE Transactions on Geoscience and Remote Sensing 58(2):982–994. https://doi.org/10.1109/TGRS.2019.2942280.

[18] Dowle M, Srinivasan A (2023) data.table: Extension of 'data.frame'. R package version 1.14.8. https://CRAN.R-project.org/package=data.table.

[19] Cui W, Wan C, Song Y (2022) Ensemble deep learning-based non-crossing quantile regression for nonparametric probabilistic forecasting of wind power generation. IEEE Transactions on Power Systems. https://doi.org/10.1109/TPWRS.2022.3202236.

[20] Efron B, Hastie T (2016) Computer Age Statistical Inference. Cambridge University Press, New York. https://doi.org/10.1017/CBO9781316576533.

[21] Fernandez-Palomino CA, Hattermann FF, Krysanova V, Lobanova A, Vega-Jácome F, Lavado W, Santini W, Aybar C, Bronstert A (2022) A novel high-resolution gridded precipitation dataset for Peruvian and Ecuadorian watersheds: Development and hydrological evaluation. Journal of Hydrometeorology 23(3):309–336. https://doi.org/10.1175/JHM-D-20-0285.1.

[22] Friedman JH (2001) Greedy function approximation: A gradient boosting machine. The Annals of Statistics 29(5):1189–1232. https://doi.org/10.1214/aos/1013203451.

[23] Gavahi K, Foroumandi E, Moradkhani H. (2023) A deep learning-based framework for multi-source precipitation fusion. Remote Sensing of Environment 295:113723. https://doi.org/10.1016/j.rse.2023.113723.

[24] Glawion L, Polz J, Kunstmann HG, Fersch B, Chwala C (2023) spateGAN: Spatio-Temporal downscaling of rainfall fields using a cGAN Approach. Earth and Space Science 10(10):e2023EA002906. https://doi.org/10.1029/2023EA002906.

[25] Gneiting T (2011) Making and evaluating point forecasts. Journal of the American Statistical Association 106(494):746–762. https://doi.org/10.1198/jasa.2011.r10138.

[26] Gneiting T, Raftery AE (2007) Strictly proper scoring rules, prediction, and estimation. Journal of the American Statistical Association 102(477):359–378. https://doi.org/10.1198/016214506000001437.

[27]     Greenwell B, Boehmke B, Cunningham J, et al. (2022) gbm: Generalized Boosted Regression Models. R package version 2.1.8.1. https://CRAN.R-project.org/package=gbm.

[28]     Hastie T, Tibshirani R, Friedman J (2009) The Elements of Statistical Learning. Springer, New York. https://doi.org/10.1007/978-0-387-84858-7.

[29]     He Y, Qin Y, Wang S, Wang X, Wang C (2019) Electricity consumption probability density forecasting method based on LASSO-Quantile Regression Neural Network. Applied Energy 233–234:565–575. https://doi.org/10.1016/j.apenergy.2018.10.061.

[30]     Hengl T, Nussbaum M, Wright MN, Heuvelink GBM, Gräler B (2018) Random forest as a generic framework for predictive modeling of spatial and spatio-temporal variables. PeerJ 6:e5518. https://doi.org/10.7717/peerj.5518.

[31]     Hollister JW (2023) elevatr: Access Elevation Data from Various APIs. R package version 0.99.0. https://CRAN.R-project.org/package=elevatr.

[32]     Hsu K-L, Gao X, Sorooshian S, Gupta HV (1997) Precipitation estimation from remotely sensed information using artificial neural networks. Journal of Applied Meteorology 36(9):1176–1190. https://doi.org/10.1175/1520-0450(1997)036<1176:PEFRSI>2.0.CO;2.

[33]     Hu Q, Li Z, Wang L, Huang Y, Wang Y, Li L (2019) Rainfall spatial estimations: A review from spatial interpolation to multi-source data merging. Water 11(3):579. https://doi.org/10.3390/w11030579.

[34]     Huffman GJ, Stocker EF, Bolvin DT, Nelkin EJ, Tan J (2019) GPM IMERG Late Precipitation L3 1 day 0.1 degree x 0.1 degree V06, Edited by Andrey Savtchenko, Greenbelt, MD, Goddard Earth Sciences Data and Information Services Center (GES DISC), Accessed: [2022-10-12], https://doi.org/10.5067/GPM/IMERGDL/DAY/06.

[35]     James G, Witten D, Hastie T, Tibshirani R (2013) An Introduction to Statistical Learning. Springer, New York. https://doi.org/10.1007/978-1-4614-7138-7.

[36]     Kang Y, Hyndman RJ, Smith-Miles K (2017) Visualising forecasting algorithm performance using time series instance spaces. International Journal of Forecasting 33(2):345–358. https://doi.org/10.1016/j.ijforecast.2016.09.004.

[37]     Kasraei B, Heung B, Saurette DD, Schmidt MG, Bulmer CE, Bethel W (2021) Quantile regression as a generic approach for estimating uncertainty of digital soil maps produced from machine-learning. Environmental Modelling and Software 144:105139. https://doi.org/10.1016/j.envsoft.2021.105139.

[38]     Ke G, Meng Q, Finley T, Wang T, Chen W, Ma W, Ye Q, Liu TY (2017) Lightgbm: A highly efficient gradient boosting decision tree. Advances in Neural Information Processing Systems 30:3146–3154.

[39]     Koenker RW (2005) Quantile regression. Cambridge University Press, Cambridge, UK.

[40]     Koenker RW (2023) quantreg: Quantile Regression. R package version 5.97. https://CRAN.R-project.org/package=quantreg.

[41]     Koenker RW, Bassett Jr G (1978). Regression quantiles. Econometrica 46(1):33–50. https://doi.org/10.2307/1913643.

[42]     Kuhn M (2023) caret: Classification and Regression Training. R package version 6.0-94. https://CRAN.R-project.org/package=caret.

[43]     Mayr A, Binder H, Gefeller O, Schmid M (2014) The evolution of boosting algorithms: From machine learning to statistical modelling. Methods of Information in Medicine 53(6):419–427. https://doi.org/10.3414/ME13-01-0122.
[44]     Meinshausen N, Ridgeway G (2006) Quantile regression forests. Journal of Machine Learning Research 7:983–999.
[45]     Nguyen P, Ombadi M, Sorooshian S, Hsu K, AghaKouchak A, Braithwaite D, Ashouri H, Rose Thorstensen A (2018) The PERSIANN family of global satellite precipitation data: A review and evaluation of products. Hydrology and Earth System Sciences 22(11):5801–5816. https://doi.org/10.5194/hess-22-5801-2018.
[46]     Nguyen P, Shearer EJ, Tran H, Ombadi M, Hayatbini N, Palacios T, Huynh P, Braithwaite D, Updegraff G, Hsu K, Kuligowski B, Logan WS, Sorooshian S (2019) The CHRS data portal, an easily accessible public repository for PERSIANN global satellite precipitation data. Scientific Data 6:180296. https://doi.org/10.1038/sdata.2018.296.
[47]     Nguyen GV, Le X-H, Van LN, Jung S, Yeon M, Lee G (2021) Application of random forest algorithm for merging multiple satellite precipitation products across South Korea. Remote Sensing 13(20):4033. https://doi.org/10.3390/rs13204033.
[48]     Papacharalampous G, Tyralis H (2022) A review of machine learning concepts and methods for addressing challenges in probabilistic hydrological post-processing and forecasting. Frontiers in Water 4:961954. https://doi.org/10.3389/frwa.2022.961954.
[49]     Papacharalampous GA, Koutsoyiannis D, Montanari A (2020) Quantification of predictive uncertainty in hydrological modelling by harnessing the wisdom of the crowd: Methodology development and investigation using toy models. Advances in Water Resources 136:103471. https://doi.org/10.1016/j.advwatres.2019.103471.
[50]     Papacharalampous GA, Tyralis H, Doulamis A, Doulamis N (2023a) Comparison of machine learning algorithms for merging gridded satellite and earth-observed precipitation data. Water 15(4):634. https://doi.org/10.3390/w15040634.
[51]     Papacharalampous GA, Tyralis H, Doulamis A, Doulamis N (2023b) Comparison of tree-based ensemble algorithms for merging satellite and earth-observed precipitation data at the daily time scale. Hydrology 10(2):50. https://doi.org/10.3390/hydrology10020050.
[52]     Papacharalampous GA, Tyralis H, Doulamis N, Doulamis A (2023c) Ensemble learning for blending gridded satellite and gauge-measured precipitation data. Remote Sensing 15(20):4912. https://doi.org/10.3390/rs15204912.
[53]     Pebesma E (2018) Simple features for R: Standardized support for spatial vector data. The R Journal 10(1):439–446. https://doi.org/10.32614/RJ-2018-009.
[54]     Pebesma E (2023) sf: Simple Features for R. R package version 1.0-13. https://CRAN.R-project.org/package=sf.
[55]     Peterson TC, Vose RS (1997) An overview of the Global Historical Climatology Network temperature database. Bulletin of the American Meteorological Society 78(12):2837–2849. https://doi.org/10.1175/1520-0477(1997)078<2837:AOOTGH>2.0.CO;2.

[56] Petropoulos F, Svetunkov I (2020) A simple combination of univariate models. International Journal of Forecasting 36(1):110–115. https://doi.org/10.1016/j.ijforecast.2019.01.006.

[57] Pierce D (2023) ncdf4: Interface to Unidata netCDF (Version 4 or Earlier) Format Data Files. R package version 1.21. https://CRAN.R-project.org/package=ncdf4.

[58] R Core Team (2023) R: A language and environment for statistical computing. R Foundation for Statistical Computing, Vienna, Austria. https://www.r-project.org.

[59] Rodrigues F, Pereira FC (2020) Beyond expectation: Deep joint mean and quantile regression for spatiotemporal problems. IEEE Transactions on Neural Networks and Learning Systems 31(12):5377–5389. https://doi.org/10.1109/TNNLS.2020.2966745.

[60] Sagi O, Rokach L (2018) Ensemble learning: A survey. Wiley Interdisciplinary Reviews: Data Mining and Knowledge Discovery 8(4):e1249. https://doi.org/10.1002/widm.1249.

[61] Sesia M, Candès EJ (2020) A comparison of some conformal quantile regression methods. Stat 9(1):e261. https://doi.org/10.1002/sta4.261.

[62] Shi Y, Ke G, Soukhavong D, Lamb J, Meng Q, Finley T, Wang T, Chen W, Ma W, Ye Q, Liu T-Y, Titov N. (2023) lightgbm: Light Gradient Boosting Machine. R package version 3.3.5. https://CRAN.R-project.org/package=lightgbm.

[63] Tareghian R, Rasmussen PF (2013) Statistical downscaling of precipitation using quantile regression. Journal of Hydrology 487:122–135. https://doi.org/10.1016/j.jhydrol.2013.02.029.

[64] Taylor JW (2000) A quantile regression neural network approach to estimating the conditional density of multiperiod returns. Journal of Forecasting 19(4):299–311. https://doi.org/10.1002/1099-131X(200007)19:4<299::AID-FOR775>3.0.CO;2-V.

[65] Tibshirani J, Athey S (2023) grf: Generalized Random Forests. R package version 2.3.1. https://cran.r-project.org/package=grf.

[66] Tyralis H, Papacharalampous G (2021a) Boosting algorithms in energy research: A systematic review. Neural Computing and Applications 33(21):14101–14117. https://doi.org/10.1007/s00521-021-05995-8.

[67] Tyralis H, Papacharalampous GA (2021b) Quantile-based hydrological modelling. Water 13(23):3420. https://doi.org/10.3390/w13233420.

[68] Tyralis H, Papacharalampous G (2023) scoringfunctions: A Collection of Scoring Functions for Assessing Point Forecasts. R package version 0.0.6. https://CRAN.R-project.org/package=scoringfunctions.

[69] Tyralis H, Papacharalampous G (2024) A review of predictive uncertainty estimation with machine learning. Artificial Intelligence Review 57(94). https://doi.org/10.1007/s10462-023-10698-8.

[70] Tyralis H, Papacharalampous GA, Doulamis N, Doulamis A (2023a) Merging satellite and gauge-measured precipitation using LightGBM with an emphasis on extreme quantiles. IEEE Journal of Selected Topics in Applied Earth Observations and Remote Sensing 16:6969–6979. https://doi.org/10.1109/JSTARS.2023.3297013.

[71] Tyralis H, Papacharalampous GA, Khatami S (2023b) Expectile-based hydrological modelling for uncertainty estimation: Life after mean. Journal of Hydrology 617(Part B):128986. https://doi.org/10.1016/j.jhydrol.2022.128986.

[72] Waldmann E (2018) Quantile regression: A short story on how and why. Statistical Modelling 18(3–4):203–218. https://doi.org/10.1177/1471082X18759142.

[73] Wang X, Hyndman RJ, Li F, Kang Y (2022) Forecast combinations: An over 50-year review. International Journal of Forecasting. https://doi.org/10.1016/j.ijforecast.2022.11.005.

[74] Weerts AH, Winsemius HC, Verkade JS (2011) Estimation of predictive hydrological uncertainty using quantile regression: Examples from the National Flood Forecasting System (England and Wales). Hydrology and Earth System Sciences 15(1):255–265. https://doi.org/10.5194/hess-15-255-2011.

[75] Wickham H (2023) tidyverse: Easily Install and Load the 'Tidyverse'. R package version 2.0.0. https://CRAN.R-project.org/package=tidyverse.

[76] Wickham H, Averick M, Bryan J, Chang W, McGowan LD, François R, Grolemund G, Hayes A, Henry L, Hester J, Kuhn M, Pedersen TL, Miller E, Bache SM, Müller K, Ooms J, Robinson D, Paige Seidel DP, Spinu V, Takahashi K, Vaughan D, Wilke C, Woo K, Yutani H (2019) Welcome to the tidyverse. Journal of Open Source Software 4(43):1686. https://doi.org/10.21105/joss.01686.

[77] Wickham H, Hester J, Chang W, Bryan J (2022) devtools: Tools to Make Developing R Packages Easier. R package version 2.4.5. https://CRAN.R-project.org/package=devtools.

[78] Wolpert DH (1992) Stacked generalization. Neural Networks 5(2):241–259. https://doi.org/10.1016/S0893-6080(05)80023-1.

[79] Wu H, Yang Q, Liu J, Wang G (2020) A spatiotemporal deep fusion model for merging satellite and gauge precipitation in China. Journal of Hydrology 584:124664. https://doi.org/10.1016/j.jhydrol.2020.124664.

[80] Xie Y (2014) knitr: A Comprehensive Tool for Reproducible Research in R. In: Stodden V, Leisch F, Peng RD (Eds) Implementing Reproducible Computational Research. Chapman and Hall/CRC.

[81] Xie Y (2015) Dynamic Documents with R and knitr, 2nd edition. Chapman and Hall/CRC.

[82] Xie Y (2023) knitr: A General-Purpose Package for Dynamic Report Generation in R. R package version 1.44. https://CRAN.R-project.org/package=knitr.

[83] Xie Y, Allaire JJ, Grolemund G (2018) R Markdown: The Definitive Guide. Chapman and Hall/CRC. ISBN 9781138359338. https://bookdown.org/yihui/rmarkdown.

[84] Xie Y, Dervieux C, Riederer E (2020) R Markdown Cookbook. Chapman and Hall/CRC. ISBN 9780367563837. https://bookdown.org/yihui/rmarkdown-cookbook.

[85] Xiong L, Li S, Tang G, Strobl J (2022) Geomorphometry and terrain analysis: Data, methods, platforms and applications. Earth-Science Reviews 233:104191. https://doi.org/10.1016/j.earscirev.2022.104191.

[86] Zhang W, Quan H, Srinivasan D (2018) Parallel and reliable probabilistic load forecasting via quantile regression forest and quantile determination. Energy 160:810–819. https://doi.org/10.1016/j.energy.2018.07.019.

[87] Zhang Y, Ye A, Nguyen P, Analui B, Sorooshian S, Hsu K (2022) QRF4P-NRT: Probabilistic post-processing of near-real-time satellite precipitation estimates using quantile regression forests. Water Resources Research 58(5):e2022WR032117. https://doi.org/10.1029/2022WR032117.